\documentclass[letterpaper, 10 pt, conference]{ieeeconf}  

\IEEEoverridecommandlockouts                              
\let\labelindent\relax
\usepackage{enumitem}
\usepackage{algorithm}
\usepackage{amsmath}
\usepackage[noend]{algpseudocode}
\usepackage[utf8]{inputenc} 
\usepackage[T1]{fontenc}    
\usepackage{microtype}      
\usepackage{graphicx}       
\usepackage{amsfonts}       
\usepackage{booktabs}       
\usepackage{nicefrac}       
\usepackage{breqn}          
\usepackage{url}            
\usepackage{xcolor}

\usepackage{subcaption}
\usepackage[font={small}]{caption}
\usepackage{siunitx}
\usepackage{breqn}
\usepackage[pagebackref=false,breaklinks=true,colorlinks,urlcolor=blue,citecolor=blue,linkcolor=blue,bookmarks=false]{hyperref}
\usepackage[capitalize, noabbrev]{cleveref}

\usepackage{color,soul}

\definecolor{chartblue}{RGB}{21, 53, 98}

\makeatletter
\let\NAT@parse\undefined
\makeatother
\usepackage[numbers]{natbib}

\usepackage{soul} 

\newcommand{\originalAS}{\mathcal{A}}
\newcommand{\latentAS}{\bar{\mathcal{A}}}
\newcommand{\originalact}{a}
\newcommand{\latentact}{\bar{a}}
\newcommand{\robotstate}{s_{\mathit{r}}}
\newcommand{\robotstatesub}[1]{s_{\mathit{r}_{[#1]}}}
\newcommand{\nonrobotstate}{s_{\mathit{nr}}}

\title{\LARGE \bf
LASER: Learning a Latent Action Space for Efficient\\ Reinforcement Learning
}

\author{
Arthur Allshire$^{*\dagger}$, %
Roberto Mart\'{i}n-Mart\'{i}n$^{*\ddagger}$, %
Charles Lin$^{\ddagger}$, %
Shawn Manuel$^{\ddagger}$, %
Silvio Savarese$^{\ddagger}$, %
Animesh Garg$^{\dagger\diamond}$
\thanks{$^{*}$ equal contribution, $^{\dagger}$University of Toronto, Vector Institute, $^{\ddagger}$Stanford University, $^{\diamond}$Nvidia \texttt{roberto.martinmartin@stanford.edu}
}%
}

\begin{document}
\maketitle

\begin{abstract}
The process of learning a manipulation task depends strongly on the action space used for exploration: posed in the incorrect action space, solving a task with reinforcement learning can be drastically inefficient. Additionally, similar tasks or instances of the same task family impose latent manifold constraints on the most effective action space: the task family can be best solved with actions in a manifold of the entire action space of the robot. 
Combining these insights we present LASER, a method to learn latent action spaces for efficient reinforcement learning. 
LASER factorizes the learning problem into two sub-problems, namely action space learning and policy learning in the new action space. 
It leverages data from similar manipulation task instances, either from an offline expert or online during policy learning, and learns from these trajectories a mapping from the original to a latent action space.
LASER is trained as a variational encoder-decoder model to map raw actions into a disentangled latent action space while maintaining action reconstruction and latent space dynamic consistency.
We evaluate LASER on two contact-rich robotic tasks in simulation, and analyze the benefit of policy learning in the generated latent action space. We show improved sample efficiency compared to the original action space from better alignment of the action space to the task space, as we observe with visualizations of the learned action space manifold. Additional details: \href{https://www.pair.toronto.edu/laser/}{\texttt{pair.toronto.edu/laser}}

\end{abstract}

\section{Introduction}
	

Deep Reinforcement Learning (RL) has fueled rapid progress in robot manipulation by enabling learning of closed loop visuomotor control policies that integrate perception and control in a single system~\cite{levine2016end}. However, the focus of end-to-end policy learning has been on the complexity of the observation (or state) space, while the decision space parameterization that affords efficient learning has been less studied. 
The best action space to learn continuous control of a robotic task depends on its specific characteristics~\cite{508440}. 

\begin{figure}[t!]
\centering
\includegraphics[width=\linewidth]{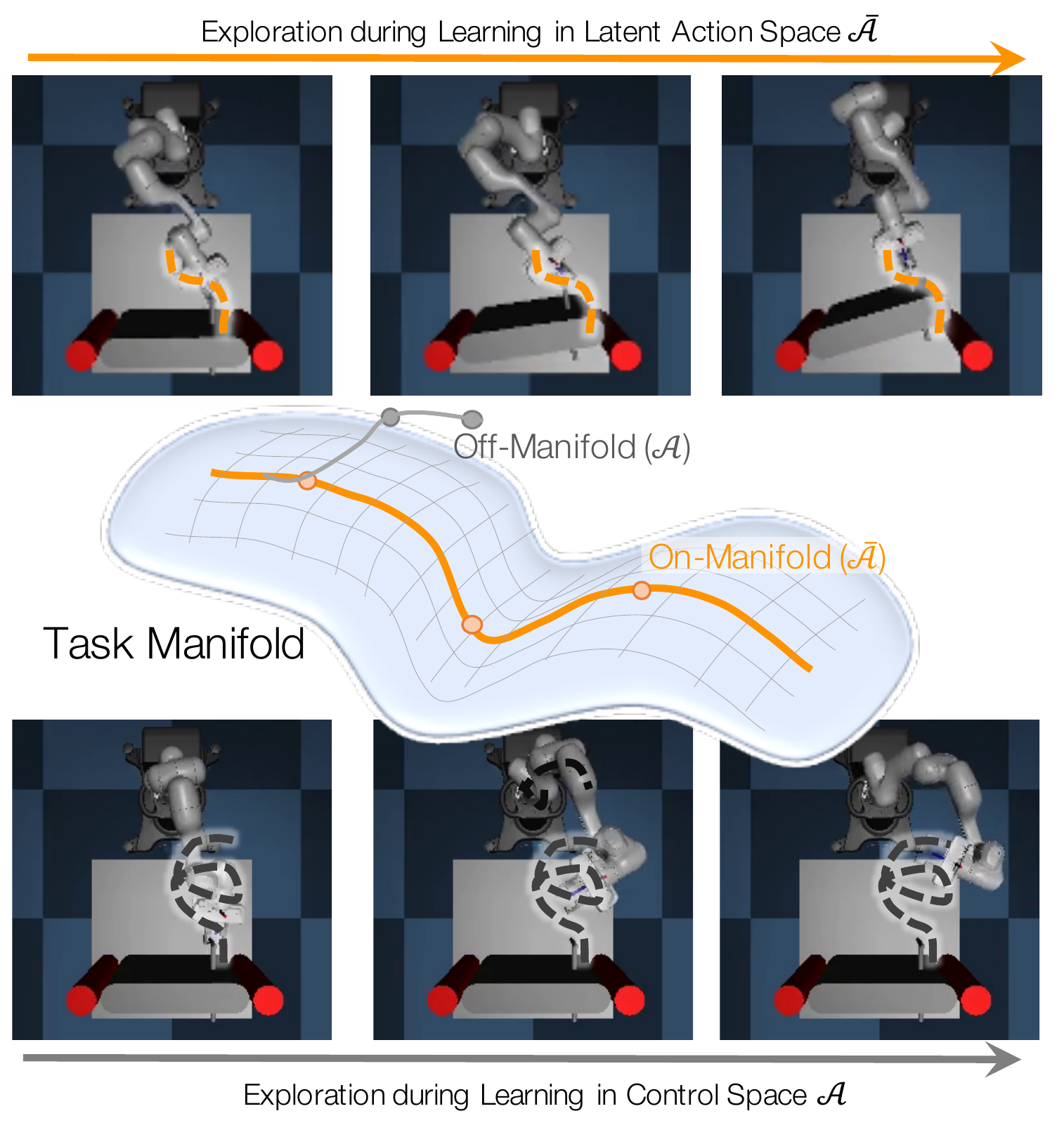}
\caption{\textbf{Learning Latent action spaces for efficient reinforcement learning}. Manipulation tasks, such as opening a door, are often structured and do not require exploration in the entire action space, only on certain manifold. LASER learns this action space manifold from data, either offline (expert) or online (training with LASER), enabling faster learning in subsequent novel instances of the task by transferring the knowledge via an efficient latent action space.}
\label{fig:loss}
\end{figure}

Consider the task of opening a door with unknown swing radius (kinematics) and torsion spring (dynamics). The agent must discover that the task progresses in a particular manifold of the action space, while yanking the arm up or down or flailing around has no value, as illustrated in Fig.~\ref{fig:loss}. 
This form of reasoning is necessary to perform many similar everyday tasks and arguably forms the basis of efficient generalization.
However, a generic RL agent’s policy is often trained in raw actuation spaces, such as joint angles or torques, discarding the latent structure in the manipulation task or task-family. 

We can model the solution of a sensorimotor control task (a policy), without loss of generality, as
a function 
$\pi(o): \mathcal{O} \rightarrow \originalAS$ which maps observations $o \in \mathcal{O}$ to control commands $\originalact \in \originalAS$ in the low-level action space that are sent to the robot's actuators.
Inspired by prior work~\cite{martin2019iros}, we propose to factorize the original problem into two sub-problems: first, learning a mapping $g(o): \mathcal{O} \rightarrow \latentAS$ from observations to actions in a new space of reference signals provided by a robot controller; and second, using the robot controller $f(\latentact): \latentAS \rightarrow \originalAS$ to map from reference signals to actuation commands. The combined control law becomes $\originalact = f\circ g(o)$, where $\latentact \in \latentAS$ is an abstract action providing a reference signal to be tracked by the mapping, $f(\cdot)$. In contrast to prior works, we propose to learn the new action space (and the controller mapping to robot's actuation commands) from the robot's experiences on similar tasks.
As a result, the original hard policy learning problem, $\pi(\cdot)$, is factorized into two coupled, simpler problems: 1) finding a suitable action representation (i.e., defining the mapping between this space and the original low-level action space, $f$), and 2) finding the mapping between observations and actions in this new latent representation space, $g$. 

In this work we propose an algorithmic approach to learn a \textit{\ul{L}atent \ul{A}ction \ul{S}pace for \ul{E}fficient exploration in \ul{R}einforcement learning} (LASER) in a data-driven manner. LASER learns from a set of task instances an optimal action space to be used in all instances of the task. Then, LASER's learned action space accelerates posterior training processes in previously seen task instances by encoding implicit structure of the task in an efficient latent action representation. 

LASER is trained as a encoder-decoder model that learns to map the manifold of low-level control inputs (in our experiments, joint positions and joint torques) to a latent action space. 
To this end, in LASER latent spaces we enforce controllability (coverage of required raw controls for a task or task family), as well as dynamical consistency across states (same action in similar states has similar effect). 
We experiment with two LASER variants: learning iteratively while policy learning (\textit{online}), and learning from batch data generated by an expert policy (\textit{offline}). We evaluate these LASER variants in two manipulation tasks and observe in both the online and offline settings that the learned action space accelerates training of new task instances.
We also analyze the learned action space and observe that the dimensions are disentangled and well aligned with the task semantics. \\

\noindent \textbf{Summary of contributions:}
\begin{enumerate}[wide, labelwidth=!, labelindent=0pt]
\item We present LASER, an algorithmic approach to learning efficient latent action spaces from off-policy or online actions of an expert to accelerate posterior training of unseen tasks, 
\item We compare learning efficiency with RL in original action space with LASER learned action space, and observe that the learned action space by LASER provides marked improvements in subsequent learning iterations, indicating a transference of information between tasks. 
\item We evaluate different variants of the LASER framework including state conditional reconstruction as well as variational reconstruction in two manipulation tasks in simulation, and find that the learned action spaces correlate clearly to the dimensions of the task space, 
\end{enumerate}

\section{Related Work}
\label{s_rw}

Robot control literature has multiple analytical maps $f(\cdot)$ (controllers) to transform action spaces (such as joint space) to task spaces (such as end effector position, velocity and acceleration)~\cite{4308708,khatib1987unified,kroger2004compliant,part1985impedance}. 
Prior work has shown that the choice of action space, often based on sophisticated analytical mappings, affects policy learning~\cite{martin2019iros,varin2019comparison} and proposed action space abstractions that facilitated learning of different families of tasks~\cite{martin2019iros,bogdanovic2019learning,gao2020learning,pervez2017learning, buchli2011learning}. 
However, the choice of an optimal action space given a task family is unclear \textit{a priori}. For instance, in a tennis swing, it is important to control position, velocity, and possibly the acceleration of the end-effector~\cite{DMP_Initial_2002}, while
in a surface-to-surface alignment task, minimizing the moment around a contact is important for robustness~\cite{khansari2016adaptive}. In this work, we propose an autonomous data-driven method to infer a latent action space better suited for learning from experiences on similar tasks.



Data-driven discovery of (near-)optimal action abstractions for efficient RL has been scarcely studied as an alternative to human-derived analytical controllers. Only the discovery of temporal action abstractions (options) have received significant attention, both in hierarchical control and option learning framework~\cite{sutton1999between,stolle2002learning,menache2002q,konidaris2011autonomous,kulkarni2016hierarchical,bacon2017option,krishnan2017transition,fang2019dynamics}. However, temporal action abstraction is orthogonal to the underlying action space and can be applied to LASER as well, and is, therefore, not the focus of this study.

Some very recent works have explored learning abstractions for action spaces. Some of them~\cite{whitney2019dynamics,van2020plannable} proposed to learn an equivalent latent full Markov decision process (MDP) of the original problem where reinforcement learning (RL) is easier. We only learn a transformation of the action space, without needing to learn the dynamics model. 
\citet{chandak2019learning} learn a continuous manifold to embed discrete actions based on the similarity of their effects. In the new continuous action space, the learning process is faster because the solution can make use of the expected correlated outcomes of close-by discrete actions. We map between two continuous action spaces for robot control.
Similarly to us, \citet{losey2019controlling} learn a new action space as a manifold within the original low-level action space. However, their method is not suited to facilitate policy learning but to simplify human teleoperation. 

The work we present here is connected to meta-learning, where the objective is to transfer knowledge from similar tasks when training on a novel instance of the task~\cite{duan2016rl,botvinick2019reinforcement,finn2017model}. These methods use a policy to transfer information between tasks; on the other hand, we propose to use a learned \textit{action space} as medium for this knowledge transfer, indicating an alternative form of meta-learning.

\section{Problem Formulation}
\label{sec:probform}

We formulate our continuous-control robot tasks as discrete-time Markov Decision Processes defined by the tuple $\mathcal{M} = (\mathcal{S},\originalAS,\mathcal{T},\mathcal{R},\gamma)$. Here, $s \in \mathcal{S} \subset \mathbb{R}^n$ is the state space, $\originalact \in \originalAS \subset \mathbb{R}^m$ is the action space, $\mathcal{T}(s'|s,\originalact)$ is the state transition model characterizing the probability of transitioning to state $s'$ from taking action $\originalact$ in state $s$, $r = \mathcal{R}(s) \in \mathbb{R}$ is a state reward function, and $\gamma \in [0,1)$ is the discount factor. The goal of a RL agent is to learn an action selection policy $\pi : \mathcal{S} \rightarrow \originalAS$ that maximizes the discounted sum of rewards from any state $s$ as $J_t = \sum_{k=0}^{\infty} \gamma^k r_t$~\cite{sutton2018reinforcement}.

Following the formalism of \citet{van2020plannable}, we assume we can lift the original MDP, $\mathcal{M} = (\mathcal{S},\originalAS,\mathcal{T},\mathcal{R},\gamma)$, into a new MDP with latent action space, $\bar{\mathcal{M}} = (\mathcal{S},\latentAS,\bar{\mathcal{T}},\mathcal{R},\gamma)$, where {$\latentact \in \latentAS$ is the latent action space} and $\bar{\mathcal{T}}(s'|s,\latentact)$ is the latent dynamics model. {We assume there exists some optimal mapping, $f: \latentAS \rightarrow \mathcal{L}$, where $\mathcal{L} \subset A$,
satisfies the property that for any task in the given task family, there exists some sequence of control actions, $a\in \mathcal{L}$, that is optimal for solving the task, so that RL exploration within the space $\mathcal{L}$ is more efficient for solving unseen tasks within the given task family compared to RL exploration in $\originalAS$. 

The latent MDP can thus be viewed as an abstraction of the original MDP such that $f$ maps latent actions, $\latentact \in \latentAS$, to control actions, $\originalact \in \originalAS$.} The insight here is that, acting within this subset of the original action space prevents exploration of actions that would never be optimal for solving those tasks. Using this insight, the goal of LASER is to learn the transformation, $f$, that maps latent actions to actions in the original action space, allowing any task within the given task family to be solved efficiently.

Assuming that LASER has found an optimal mapping, $f$, from the latent action space to the original action space, a RL policy $\pi: S\rightarrow \latentAS$ would be able to explore the optimal region of the original action space by acting in the latent action space of the lifted latent MDP $\latentAS$. A policy on the latent MDP, $\pi: s\mapsto \latentact$, generates a policy on the original MDP, $\pi': s\mapsto f(\latentact)$. As shown by \citet{van2020plannable}, the generated policy on the original MDP is optimal if the policy on the lifted MDP is optimal.

\section{Learning Action Spaces for Efficient RL}
\label{sec:learning}

\begin{figure}[t!]
\begin{center}
\includegraphics[width=\linewidth]{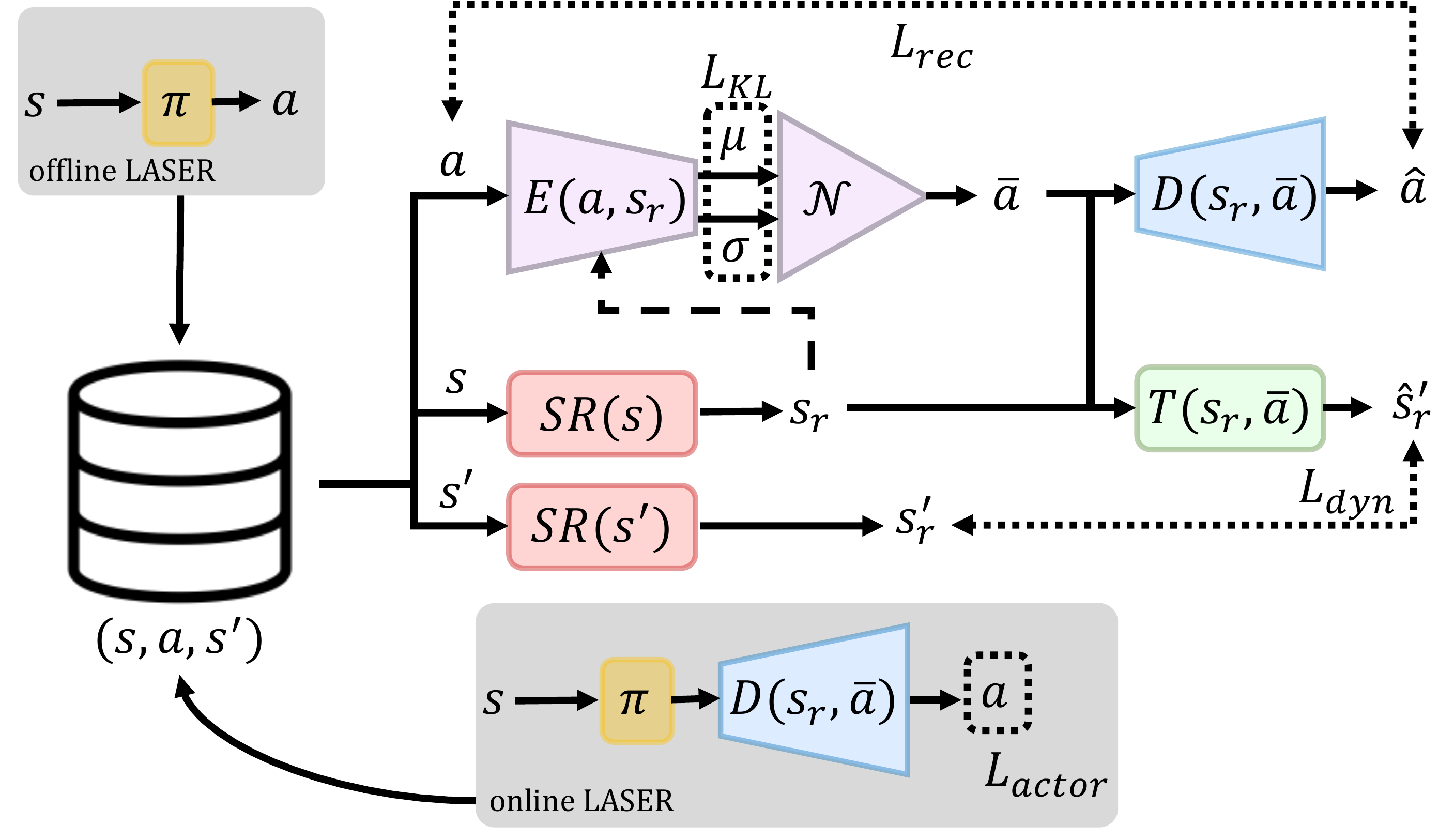}
\caption{
\textbf{LASER Overview:} We train a latent action space $\mathcal{\bar{A}}$ using batches of tuples $(s,\originalact,s')$. These batches come from a previously generated dataset (LASER \textbf{offline}) or from a dynamically generated replay buffer (LASER \textbf{online}). Actions are used in an encoder-decoder architecture with reconstruction loss, $L_{\mathit{rec}}$. The encoder's Gaussian prediction, $(\mu,\sigma)$, is regularized via the KL loss $L_{\mathit{KL}}$. The robot components of state, $s$, and next state, $s'$, and the latent actions are used to train a latent state transition model, $T$, with a dynamics loss, $L_{\mathit{dyn}}$. In the LASER online variant, this process alternates with policy learning that generates new tuples to add to the dataset (replay buffer), and uses the actor loss, $L_{\mathit{actor}}$. The policy generates actions in the latent action space, $\mathcal{\bar{A}}$, that are decoded into robot's original action space. 
Gradients updates from the actor loss propagate also through and are applied to online LASER's decoder during policy updates. 
The learned LASER action space $\mathcal{\bar{A}}$ accelerates subsequent training for new instances of the task.}
\label{fig:sys}
\end{center}
\end{figure}

This section describes LASER, our algorithm for representation learning of latent action spaces (see Fig.~\ref{fig:sys}). LASER learns a transformation between a latent action space, $\latentAS$, and original action space, $\originalAS$, acting as a latent controller for more efficient policy learning in the latent action space. We outline the learning LASER algorithm in online and offline settings, and a procedure for transfer learning with LASER. 



\subsection{Representing Latent Action Spaces}



LASER learns a MDP transformation map between original actions $\originalact$ and latent actions $\latentact$, as presented in Sec.~\ref{sec:probform}.
We assume that the mapping between latent and original action spaces depends on the current state of the robot. This is the case for the analytical robot controllers that we take as inspiration for this work~\cite{martin2019iros,siciliano2016springer}, as we can see with an example. Suppose that the original action space $\originalAS$ of the MDP is the space of torques at the joints of a robot, a frequent low-level action space for robots in research. Moreover, suppose that the task is best learned in a latent action space $\latentAS$ corresponding to desired positions for each robot's joint. Given an action in the latent space, $\latentact \in \latentAS$, the joint torques corresponding to the desired joint position would depend \textit{on the current state of the robot}: if the current robot state were close to or at the desired joint position $\bar{a}$, the torques would be close to zero, but if the robot state were very different to the desired joint positions the transformation into the original action space would lead to larger joint torques, $a$. The transformation is, however, independent of other state information such as the state of the environment or information about the task; this information is used by the policy to deciding what actions to perform.

Based on these insights, in LASER we propose to learn a representation mapping between original and latent action spaces with a encoder-decoder neural architecture conditioned on the current state of the robot. We assume that the state of the environment, $s$, can be separated into a distinct robot state $\robotstate$ and non-robot state $\nonrobotstate$: $s = [\robotstate, \nonrobotstate]$. $\robotstate$ can be extracted from the full state ($\robotstate = \mathit{SR}(s)$) and contains the kinematic and dynamic information of the current state of the robotic agent such as joint configurations, accelerations and Cartesian pose of the end-effector, provided by robot's proprioceptive sensing. $\nonrobotstate$ contains other information about the state of the environment and that can be task-relevant such as the goal of the task. 

The action encoder of LASER is a variational neural network $E_{\theta_E} : \originalAS \times \mathcal{S}_{\mathit{r}} \rightarrow \latentAS$ parameterized by $\theta_E$ that encodes an action $\originalact$ in the original action space $\originalAS$, conditioned on the current robot state $\robotstate$, into a latent action $\latentact \sim E_{\theta_E}(\originalact,\robotstate)$. The function $f: \latentAS\rightarrow\originalAS$ defined in Sec.~\ref{sec:probform} for mapping from latent actions to control inputs in the original action space will be represented by a latent state-dependent variational decoder neural network, $D_{\theta_D} : \mathcal{S}_{\mathit{r}} \times \latentAS \rightarrow \originalAS$ parameterized by $\theta_D$, where $\hat{\originalact} = D_{\theta_D}(\robotstate,\latentact)$ is the reconstruction of $a$, an action in the original space that would have resulted in $\latentact \sim E_{\theta_E}(\originalact)$. Finally, the latent state transition function $\bar{\mathcal{T}}_{\mathit{r}}(\robotstate'|\robotstate,\latentact)$ can be modeled as a function $T_{\theta_T} : \mathcal{S}_{\mathit{r}} \times \latentAS \rightarrow \mathcal{S}_{\mathit{r}}$ to output the next robot state $\robotstate' = T_{\theta_T}(\robotstate, \latentact)$.

After learning an action space representation with LASER, an RL policy $\pi : \mathcal{S} \rightarrow \latentAS$ can then be trained in this latent action space using the decoder to map the policy's latent actions back to the original action space $\originalact = {D}_{\theta_D}(\robotstate,\latentact)$.

\subsection{Learning an Action Representation with LASER}
\label{learning-action-repr}

To learn a latent action space with LASER defined by an encoder-decoder action space mapping, we will leverage a replay buffer with experiences of attempts to complete the task. These demonstrations may be collected ahead of time by an expert policy as in offline LASER, or using a suboptimal policy and collected as training progresses, as in online LASER (see Sec.~\ref{subsec:variants}). The replay buffer contains triplets of state, action, next state, $(s, \originalact, s')$, from the original MDP with $\originalact \in \originalAS$. Therefore, the dataset represents a distribution of control actions in the associated state for useful for achieving tasks within the intended task family. 


\begin{figure}[t!]
\begin{center}
\begin{subfigure}{0.23\textwidth}
\includegraphics[width=0.99\linewidth]{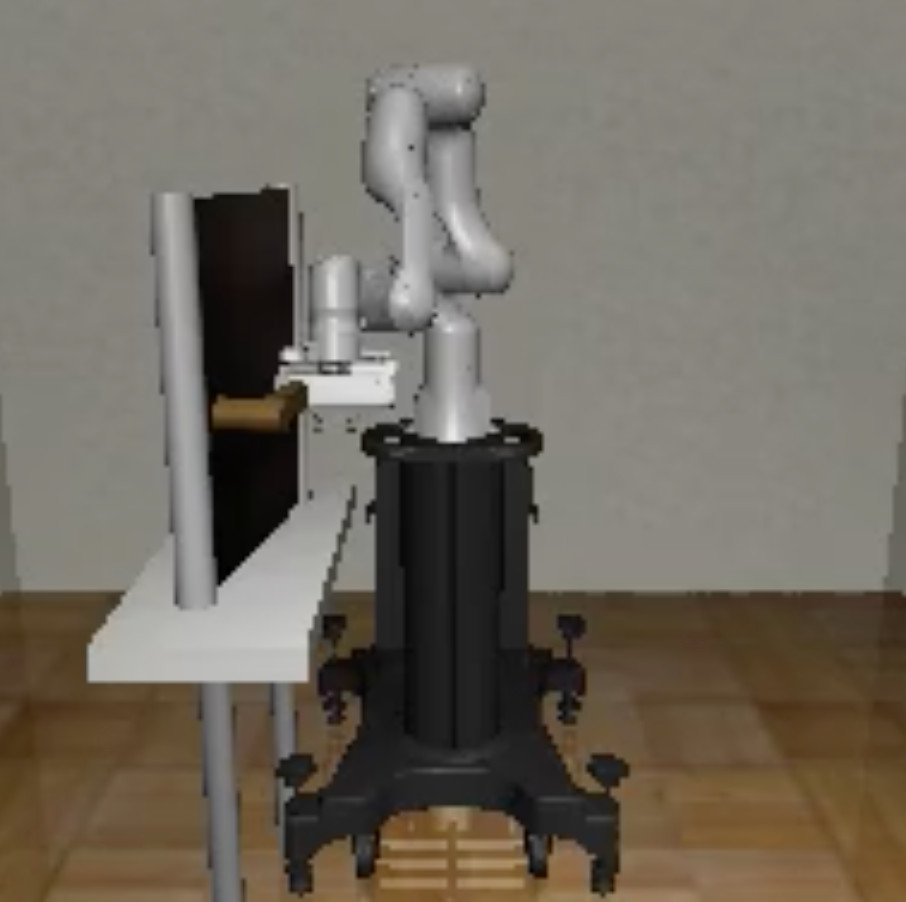}
\caption{\texttt{\footnotesize Door}}
\label{f:pe}
\end{subfigure}
\hfill
\hfill
\begin{subfigure}{0.23\textwidth}
\includegraphics[width=0.99\linewidth]{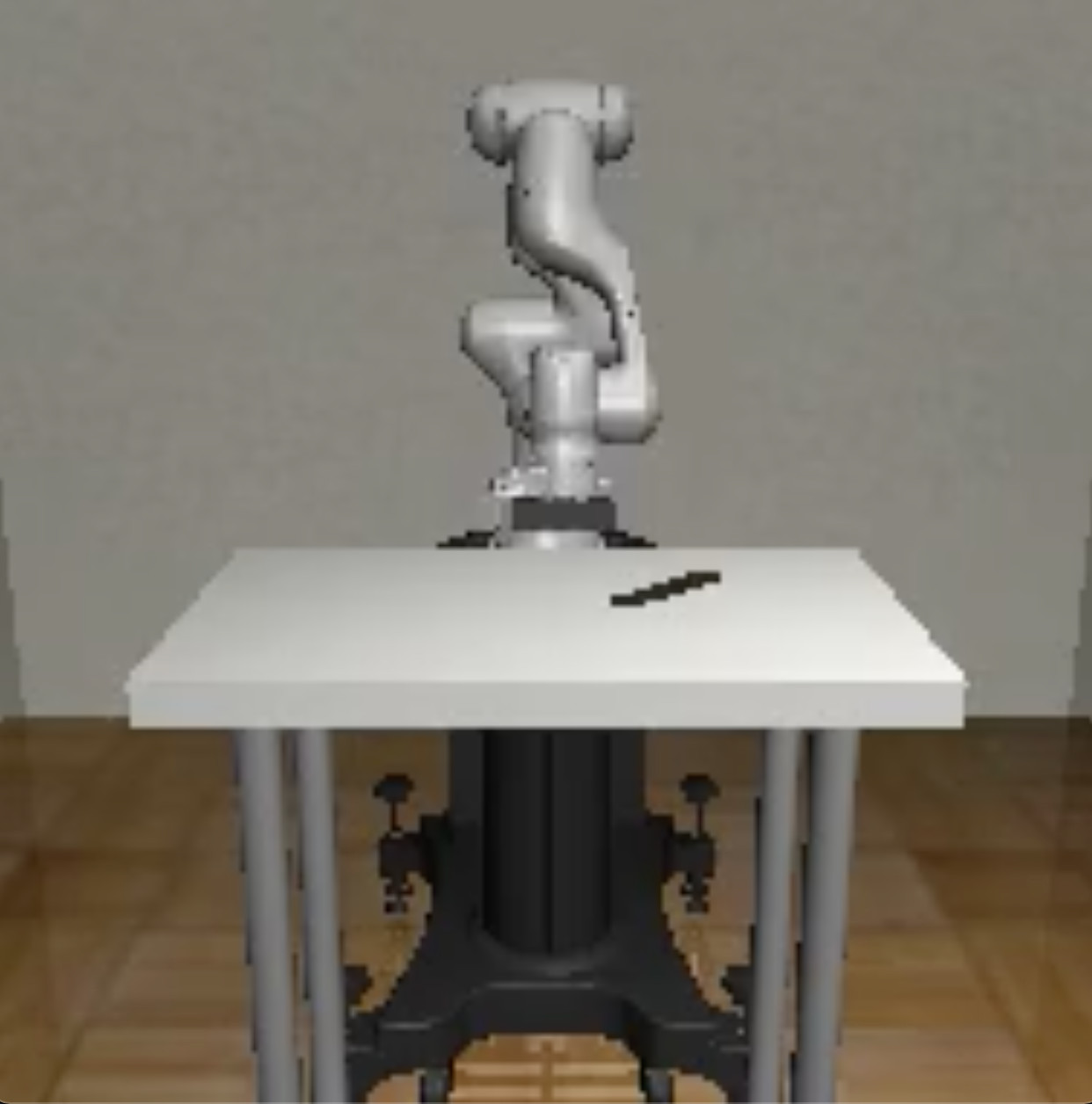}
\caption{\texttt{\footnotesize Wipe}}
\label{f:pf}
\end{subfigure}
\end{center}
\caption{\textbf{Two simulation tasks used to evaluate LASER.} a) \texttt{Door}: the agent controls a Panda robot and has to open a door a given angle.
b) \texttt{Wipe}: the agent controls a Panda robot with an eraser end-effector and needs to contact a surface and wipe the dirt elements on it. The two tasks involve solving problems in clear contact-generated submanifolds of the action space. LASER can help an agent learn the actions to traverse these manifolds and accelerate training in new instances of the tasks.}
\label{fig:robots}
\end{figure}

As shown in Fig.~\ref{fig:sys}, the LASER framework incorporates several loss terms to find a suitable latent action space.
In order to allow for a latent action space to have lower dimensionality than the original action space, we use autoencoders that preserve the principal dimensions of variation of the original space in the latent action space. Thus, the first loss we impose is a \textbf{reconstruction loss}. The decoder $D_{\theta_D}$ will be trained to reconstruct an action $\hat{\originalact} = D(\robotstate,\latentact;\theta_D)$ from the latent action $\latentact \sim E(\originalact,\robotstate;\theta_E)$ from the encoder and the given robot state $\robotstate$. This results in a typical autoencoder reconstruction loss~\citep{baldi2012autoencoders} defined as:
\begin{equation}
    L_{\mathit{rec}}(s,\originalact,\theta_E,\theta_D) = \| \originalact - D(\robotstate,E(\originalact,\robotstate;\theta_E);\theta_D) \|_2^2
\end{equation}

As found in previous work on learning latent action spaces~\cite{losey2019controlling}, an effective latent action space should satisfy three properties: latent controllability, latent consistency, and latent scaling. \textit{Latent controllability} requires the dynamics transitions between two consecutive latent states $\robotstate$ and $\robotstate'$ to mimic the transition between their corresponding states in the original MDP, $s$ and $s'$.
\textit{Latent consistency} enforces similar state transition behavior when the same latent action is taken in similar states. Assuming that executing a latent action $\bar{a}_1$ at the state $\robotstate$ results in transition to state $\robotstatesub{1}'$, if we execute another latent action $\bar{a}_2$ close to $\bar{a}_1$ ($|\bar{a}_1 - \bar{a}_2| < \delta_{\bar{a}}$) at $\robotstate$, we transition to a new state $\robotstatesub{2}'$ that is close to $\robotstatesub{1}'$ ($|\robotstatesub{1}' - \robotstatesub{2}'| < \delta_\robotstate$). Finally, \textit{latent scaling} ensures that applying larger latent actions leads to larger changes in latent state. As found by \citet{van2020plannable}, these properties can be achieved by incorporating a \textbf{latent state dynamics loss} that forces the predicted state from the learned latent state transition model, $\hat{s}'_{\mathit{r}} = T(\robotstate,\latentact;\theta_T)$ to be close to the true next latent state $\robotstate'$ for a latent action $\latentact \sim E(\originalact;\theta_E)$:

{\small
\begin{equation}
      L_{\mathit{dyn}}(s,\originalact,s',\theta_E,\theta_T)
      = \| \robotstate' - T(\robotstate,E(\originalact,\robotstate;\theta_E);\theta_T) \|_2^2 
\end{equation}
}
Finally, we also include a \textbf{regularization component} to the loss in the form of a Kullback–Leibler (KL) divergence term, as is common in variational autoencoder architectures~\citep{kingma2013auto}. The KL loss ensures the encoder learns a smooth latent space distribution with zero mean:
\vspace{-13pt}

\begin{equation}
    L_{\mathit{KL}}(\originalact,\theta_E) = \mathit{KL}(\mathcal{N}(\mu(\originalact;\theta_E),\sigma(\originalact;\theta_E))\ \| \ \mathcal{N}(0,1))
\end{equation}

The final LASER loss function is a weighted-sum of the aforementioned losses with a unique constant weighting for each component. The weights allow to prioritize some objectives over others, e.g. the reconstruction over the KL divergence loss, following a $\beta$-VAE approach~\citep{higgins2017beta}:

{\small
\begin{dmath}
L(s,\originalact,s',\theta_E,\theta_D,\theta_T) =  \beta_{\mathit{rec}}L_{\mathit{rec}} +  \beta_{\mathit{dyn}}L_{\mathit{dyn}} +  \beta_{\mathit{KL}}L_{\mathit{KL}}
\label{eqloss}
\end{dmath}
}

\begin{figure}[t]
    \centering
    \def \reswidth {1.0}
    \includegraphics[width=\reswidth\linewidth]{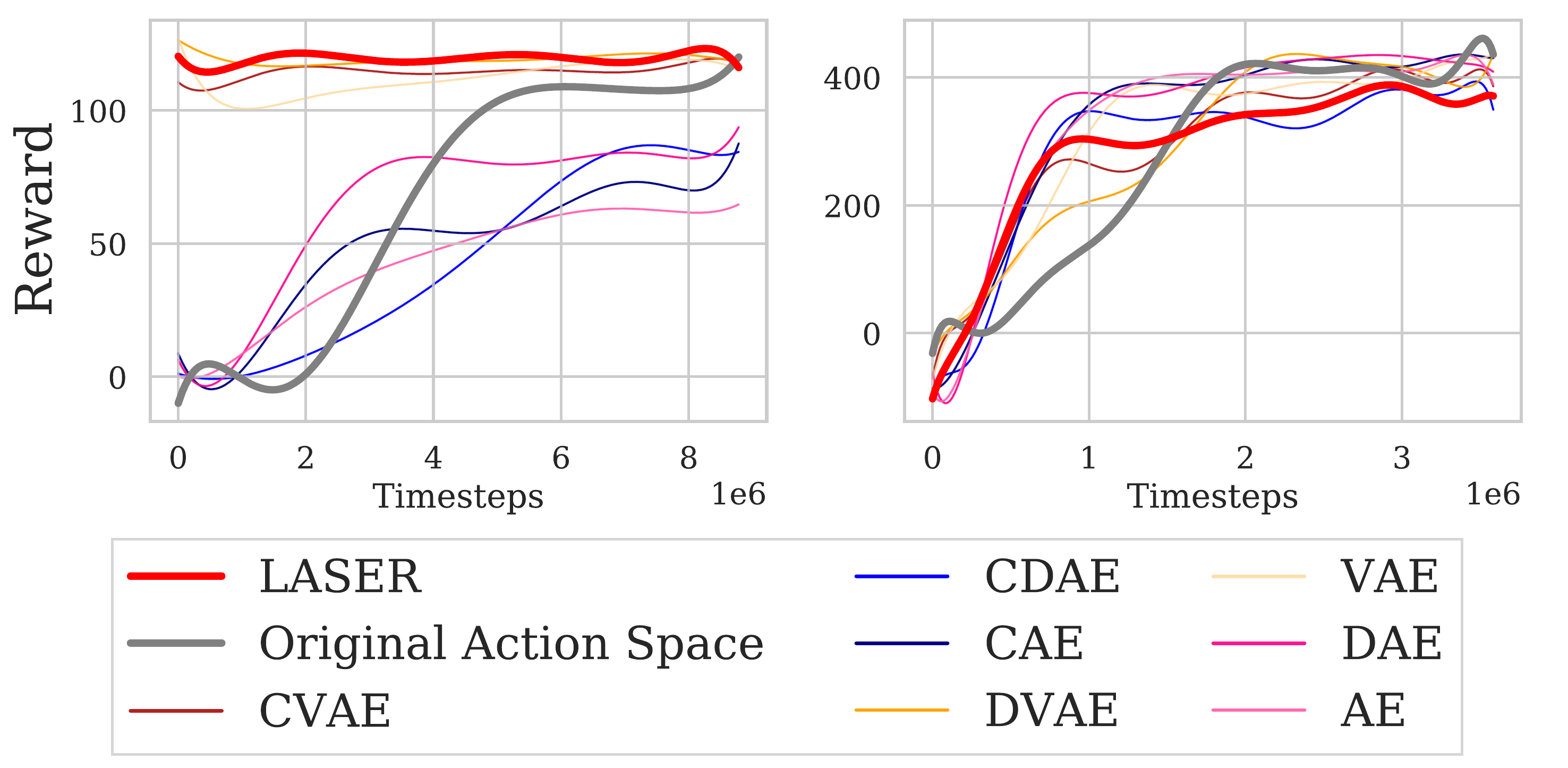}
    \caption{\textbf{Exp1. LASER trained on offline batch data:} SAC on the original action space, a LASER action space, and on ablations of LASER (Sec.~\ref{subsec:expts}) on the \texttt{Door} (\textit{left}) and \texttt{Wipe} (\textit{right}) tasks. LASER and LASER ablations are trained offline (Sec.~\ref{subsec:variants}) on trajectories sampled from an expert SAC. Our results show that training in the LASER action space converges faster than training in the original action space.}
    \label{fig:offline-results}
\end{figure}

\subsection{Offline and Online LASER Variants}
\label{subsec:variants}

There are two alternative variants to train LASER: offline and online (Fig.~\ref{fig:sys}). Both train using the LASER loss of Equation~\ref{eqloss}, but differ in their training process. Also, the online variant leverages an additional loss, as we will see below. In the \textbf{offline LASER variant}, we leverage a dataset of expert policy experiences acquired for a \textit{base task} in order to improve learning efficiency on subsequent \textit{transfer tasks}. The offline LASER process is as follows. First, we train a standard RL algorithm to convergence on the base task, using the original action space of the robot. After convergence, we roll-out episodes using the trained policy and generate a dataset of experiences (consisting of state, action, next state tuples) and train LASER on this dataset. Finally, we train RL agents on transfer tasks using the learned LASER action space: the trained policies will generate actions in the latent space that will be transformed into the original action space of the task by the trained LASER's state-conditioned decoder.

In the \textbf{online LASER variant}, we train LASER at the same time that a policy is learning a task for without any pre-training of either the policy or the representation. The policy is using the non-stationary action space provided by LASER. We alternate between LASER training using experiences from the replay buffer of the policy, and policy training in the latest LASER space. An online LASER policy consists of an multi-layer perceptron head followed by the LASER action decoder (Fig.~\ref{fig:sys}, bottom). The decoder shares the weights with the decoder in the encoder-decoder LASER architecture. To improve LASER training, we exploit the gradients from the policy to optimize LASER's decoder: we propagate gradients through the decoder during policy training iterations. We found that this process significantly speeds up online training. In Sec.~\ref{subsec:expts} (Exp 2), we show that online LASER procedure can learn the action representation and the policy simultaneously, without incurring any efficiency penalty when compared to training in the original action space. Moreover, it retains the benefit of being able to learn new policies in the learned action space.
\section{Experimental Evaluation}\label{sec:result}

In our experiments, we aim to answer four questions:
    
\begin{enumerate}[wide, labelwidth=!, labelindent=0pt]
    \item How does a RL policy learning on LASER action space, $\mathcal{\bar{A}}$, trained on offline batched data, compare to a RL policy learning in the original action space, $\mathcal{A}$?
    \item How does the action space learned with LASER transfer to different variants of the task?
    \item Does online learning of a LASER action space affect policy learning?
    \item How do LASER action spaces align with the natural dimensions of a task?
\end{enumerate}

\begin{figure}[t]
    \centering
    \def \reswidth {1.0}
    \includegraphics[width=\reswidth\linewidth]{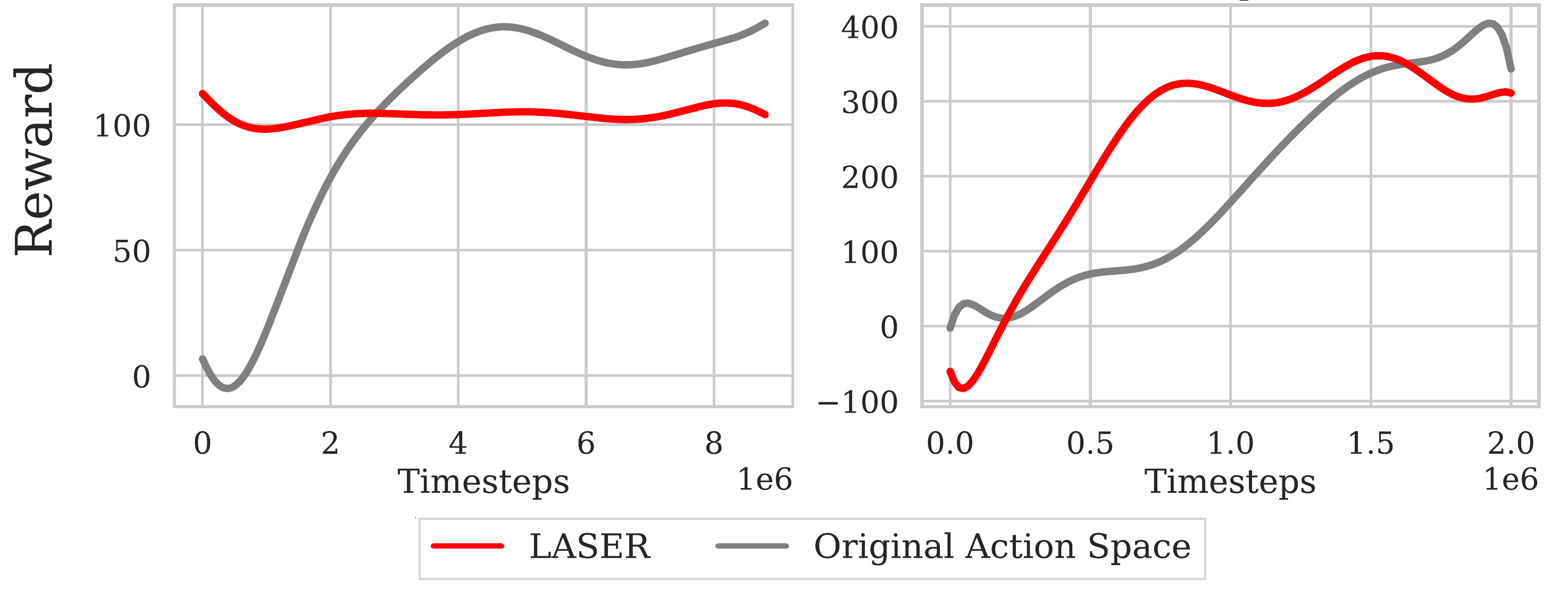}
    \caption{\textbf{Exp2. LASER task transfer:} SAC on the original and LASER action spaces in new instances of \texttt{Door} (\textit{left}), and \texttt{Wipe} (\textit{right}). In both tasks, we observe a significant benefit in efficiency when learning in the LASER action space compared to the original action space, indicating a transfer of information in the form of an efficient action space.}
    \label{fig:crosstask-results}
\end{figure}

\subsection{Experimental Setup}
\label{setup}
\noindent \textbf{Environments:}
We conduct experiments on two contact-rich tasks using the RoboSuite simulator ~\cite{robosuite2020}: \texttt{Door} and \texttt{Wipe} (Fig.~\ref{fig:robots}). The goals are to grasp and move a door to a predefined configuration, and to wipe spots of ``dirt'' on a table, respectively. All experiments are conducted using 16 environments in parallel. We use Soft Actor Critic (SAC,~\citep{haarnoja2018soft}) as our reinforcement learning algorithm. In the \texttt{Door} environment, the original action space is joint torque control; in the \texttt{Wipe} environment, the original action space is joint position control.

\noindent \textbf{Ablations:}
LASER involves learning three models: 1) a robot state--conditioned encoder network, $E_{\theta_E}(\originalact,\robotstate)$, for lifting actions from the original action space, $\originalAS$, to the latent action space, $\latentAS$; 2) a decoder network $D_{\theta_D}(\robotstate,\latentact)$ for mapping actions from the latent action space, $\latentAS$, back to the original action space, $\originalAS$; and 3) a latent state transition function $T(\robotstate,\latentact;\theta_T)$ to impose a smooth transition of robot states from applying latent actions in the latent MDP. LASER uses a state-conditioned decoder (``C''), and losses in dynamics (``D'') and KL to an isotropic normal distribution in latent action space, corresponding to a variational auto-encoder architecture (``VAE''), summarized as the model ``CDVAE''. We compare LASER to the following ablations: 1) variants without state-conditioning at the encoder (no ``C'' in name), 2) variants without state-conditioning at the encoder (no ``D'' in name), and 3) variants without KL regularization leading to simple auto-encoder architecture (``AE'' name instead of ``VAE''). For all, we use $\beta_{\mathit{rec}}=1$, $\beta_{\mathit{dyn}}=1$, $\beta_{KL} = 0$ for non-variational variants, and $\beta_{KL} = 0.01$ for variational ones (Eq.~\ref{eqloss}). 

\begin{figure}[t]
\centering
\def \reswidth {1.0}
\includegraphics[width=\reswidth\linewidth]{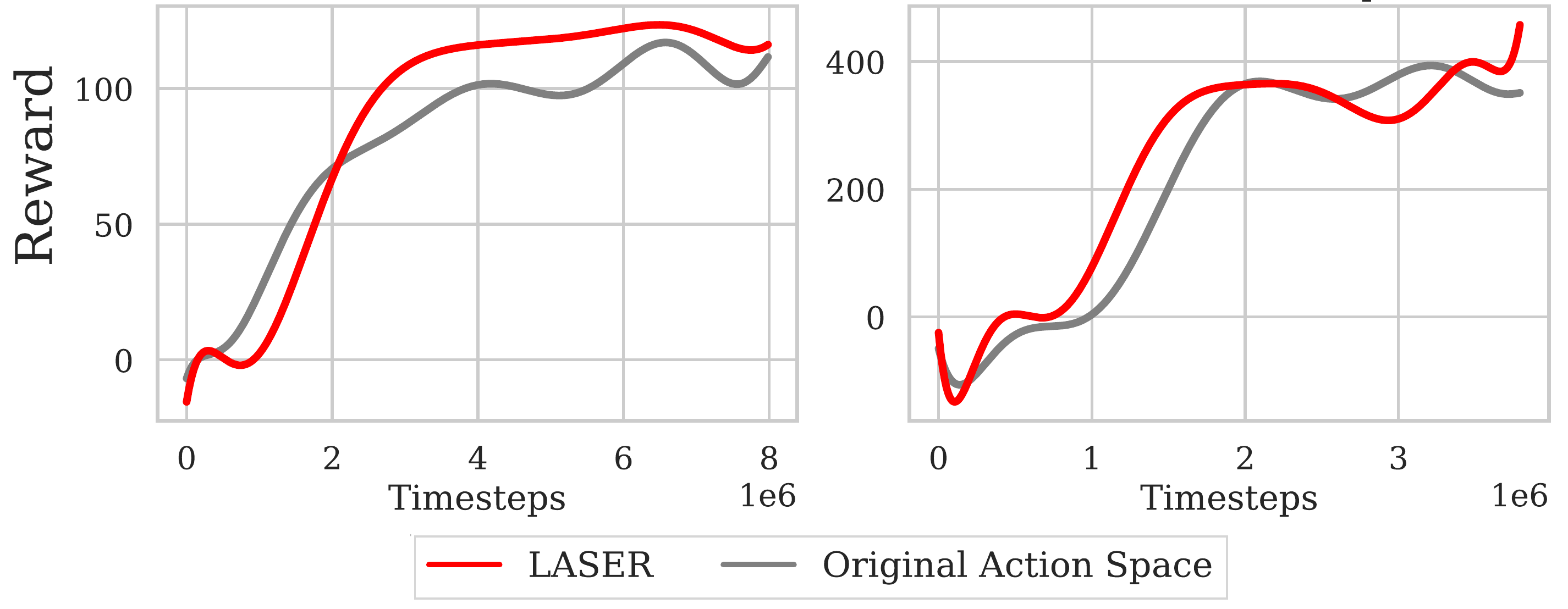}
\caption{\textbf{Exp3. LASER trained online with policy learning:} SAC on the original and online LASER action spaces on \texttt{Door} (\textit{left}), and \texttt{Wipe} (\textit{right}). The improved performance (faster and more optimal convergence) of SAC in the LASER action spaces indicates that it is possible to learn a latent action space simultaneously with policy learning, and that learning such an action space improves the efficiency of policy learning.}
\label{fig:onlineresults}
\end{figure}

\subsection{Experiments}
\label{subsec:expts}

\noindent \textbf{Exp1. LASER trained on offline batch data:}
The aim of this experiment is to test whether policy learning is more efficient in a learned latent action space than in an original action space. First, we train a RL policy to convergence on a set of tasks. We then sample 1,000 episodes from this expert policy on each task to form a dataset of expert experiences, which we use to train SAC on action spaces learned by LASER and its ablations.

The results are shown in Fig.~\ref{fig:offline-results}.
In both the \texttt{Door} and \texttt{Wipe} tasks, SAC with offline LASER converges faster than SAC on the original action space. In \texttt{Door}, SAC achieves and maintains a reward of over 100 zero-shot in the LASER action space, while it takes 500,000 steps to achieve the same reward in the original action space. In the \texttt{Wipe} task, SAC reaches a reward of 300 in 600,000 steps with LASER, compared to in 1.5 million in the original action space. This suggests that LASER is able to learn action spaces that simplify learning contact-rich tasks with manipulation of constrained mechanisms.

\noindent \textbf{Exp2. LASER task transfer:} We investigate task transfer to both the offline version of LASER presented in Sec.~\ref{subsec:variants}. We use the action space learned with LASER offline for the \texttt{Door} and \texttt{Wipe} tasks (\textbf{Exp1}) to learn with SAC in a different instance of the base task and compare to learning in the original action space. 

In the unseen \texttt{Door} transfer task, we increase the damping coefficient of the door by a factor of 5. The optimal joint torques to solve the transfer task are of higher magnitude than the optimal joint torques to solve the base task, and hence the optimal submanifold of the original joint torque action space is out of the distribution of the LASER action space. In the unseen \texttt{Wipe} transfer task, we have the robot wipe randomly placed circles (instead of lines), so it needs to learn a different motions within the same task manifold. 

In \texttt{Door} variant, we reach a reward of 100 zero-shot in the LASER action space, compared to in over 200,000 steps in the original action space. Because the optimal task manifold of the transfer task differs from that of the base task, SAC in the LASER action space converges slightly less optimally than SAC in the original action space does. However, the zero-shot performance of SAC with offline LASER suggests that the action space captures information common to solving both tasks. In the \texttt{Wipe} variant, SAC reaches a reward of 300 in 600,000 steps with LASER, only achieving the same performance in 1.4 million steps in the original action space.

Our results, depicted in Fig.~\ref{fig:crosstask-results}, suggest that the action space learned by LASER provides a useful representation for learning new task instances unseen during representation learning, improving efficiency in subsequent training processes.


\noindent \textbf{Exp3. LASER trained online with policy learning:} In this experiment we benchmark the performance when interleaving representation learning (using LASER losses) with policy learning (using the SAC losses), as described in Sec.~\ref{subsec:variants}. Surprisingly, SAC in the action space learned simultaneously with online LASER has converges faster than SAC in the original action space, as shown in the training curve in Fig.~\ref{fig:onlineresults}. This suggests that online learning of action space does not impede policy learning but rather facilitates it.


\begin{figure}[t]
\centering
\def \reswidth {1.0}
\includegraphics[width=0.85\linewidth]{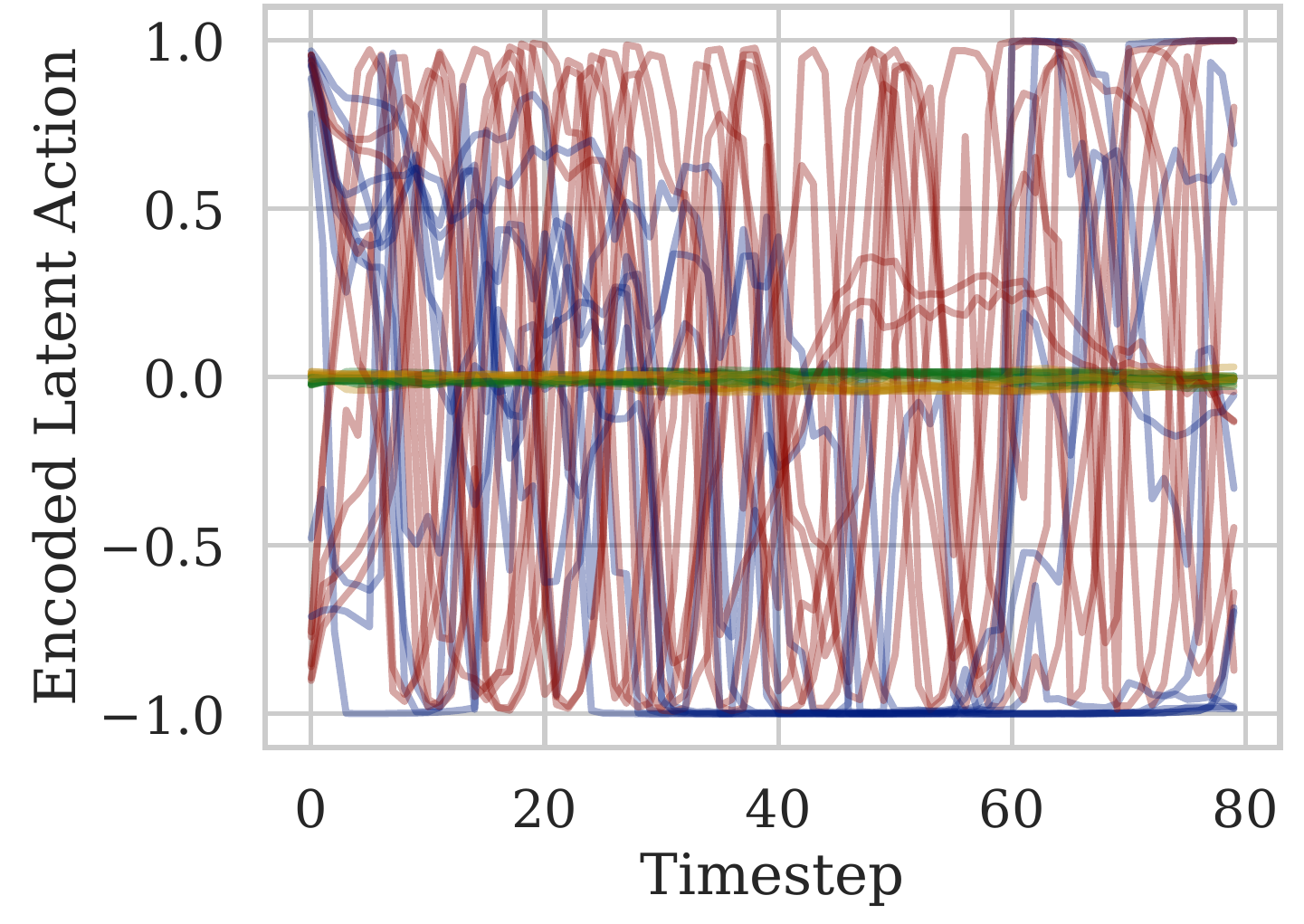}
\vspace{0pt}
\caption{
\textbf{Exp4a. Dimensionality analysis (best viewed in color):} Mean of each dimension of the variational encoder output for 10 different rollouts; each latent action dimension is shown in a different colour. Only 2 out of the 4 dimensions are utilised (those coloured in blue and red; while green and orange are zero), showing that LASER recovers an action space that is as low-dimensional as possible for efficient policy learning. 
}
\label{fig:encodedactions}
\end{figure}

\vspace{2pt}
\noindent \textbf{Exp4. Qualitative action space evaluation:} In this experiment, we investigate the action space manifold learned by LASER in the \texttt{Wipe} environment. Our goal is to observe whether the latent action space aligns with the natural dimensions of the task space in the \texttt{Wipe} task.

In a first experiment, we retrieve the encoded actions during rollouts of the \texttt{Wipe} task to inspect the dimensionality of the learned latent manifold. The results are depicted in Fig.~\ref{fig:encodedactions}. We observe that although we allow a latent action space of dimensionality 4 to be learned, LASER reduces the dimensionality of the latent space to 2 dimensions, learning a manifold that corresponds to the 2-dimensional task-space.

\begin{figure}[t!]
\centering
\includegraphics[width=\linewidth,
trim=1.5cm 0.5cm 0cm 2cm, clip]{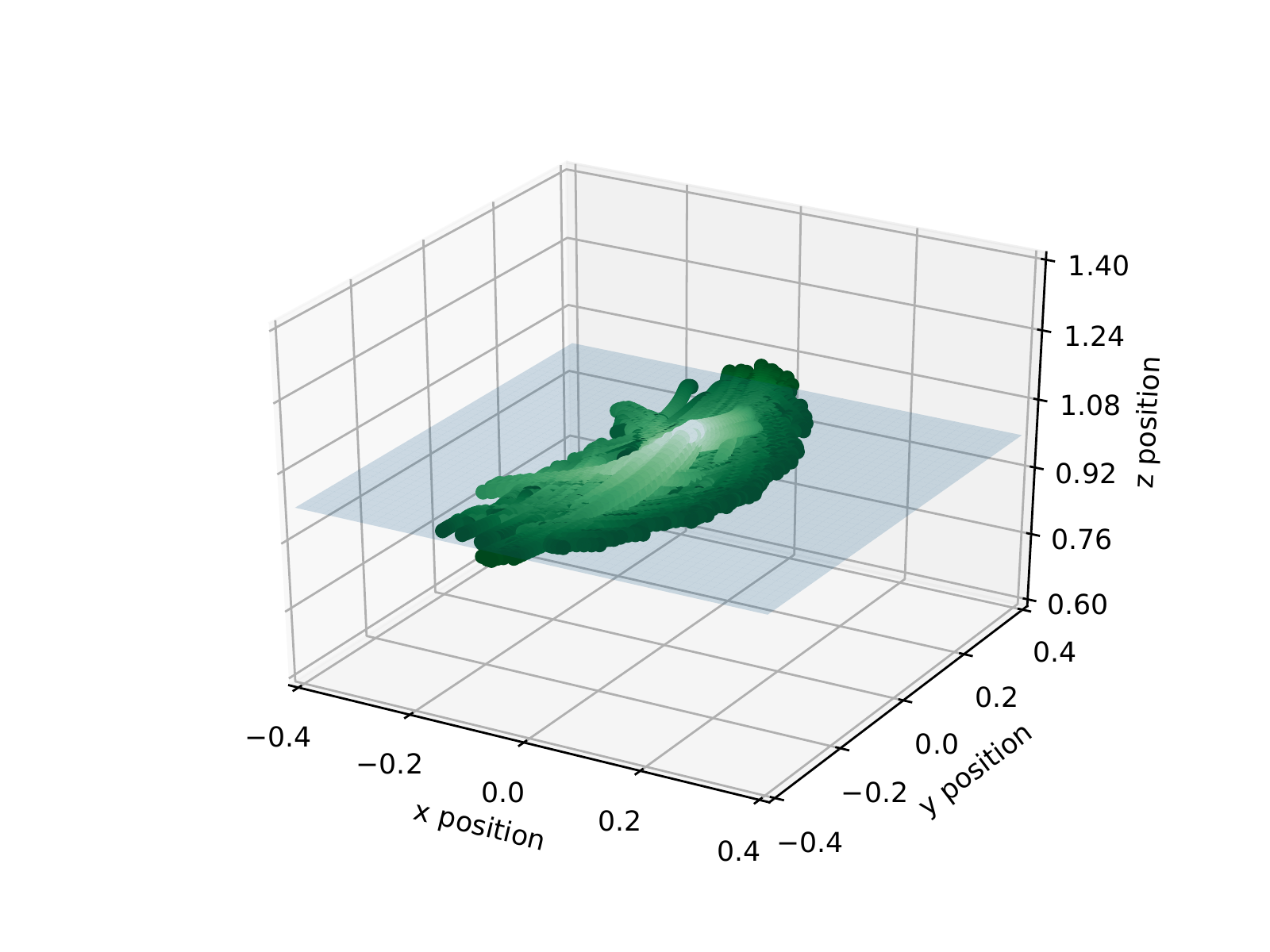}
\vspace{-10pt}
\caption{\textbf{Exp4b. Latent space traversal (best viewed in color):} End-effector positions from 200 trajectories within a LASER space on \texttt{Wipe}. The traversals lead to end-effector motions approximately parallel to the $xy$-plane. The blue plane represents the table in the \texttt{Wipe} task. Trajectories are colored by timestep in rollout, with lighter colors representing earlier timesteps. LASER learns a space which aligns with the natural task space.}
\label{fig:wipinglatent}
\end{figure}

In a second experiment, we traverse the latent action space by continuously applying latent actions in a sinusoidal pattern, executing actions obtained from the LASER decoder outputs, and collecting the end-effector position. The end-effector motion is depicted in Fig.~\ref{fig:wipinglatent}.
We observe that the robot is controlled along the dimensions of interest to the task, the $xy$-plane over the table surface. The end-effector's motion is primarily lateral, encapsulating the motions necessary to wipe and pan around the table.
These experiments indicate that the learned action representation aligns well with the task subspace, mapping to the submanifold of the original action space on which the task should be executed.


\section{Conclusion}\label{sec:conclusion}

We presented LASER, an approach to learning a latent action space for efficient reinforcement learning. LASER transforms the original MDP of an RL problem into a new MDP, where exploration is easier. The action representation is learned from expert data in an offline (pre-acquired data) or online manner (while the data is acquired). LASER is a variational encoder-decoder model that maps actions in the original action space into a disentangled latent space while maintaining both state-conditioned reconstruction as well as latent space dynamic consistency.
We evaluated LASER and LASER ablations in two contact rich manipulation tasks (door opening, and surface wiping) and combined state-of-the-art policy learning algorithms (SAC). 
Our results revealed that LASER often facilitates training in the same tasks and helps transfer knowledge for faster exploration and convergence in transfer tasks. Visualizations of the learned action space indicate that LASER learns an action space aligned with the natural dimensions of the task-space, leading to the observed improvement in subsequent training processes.



\clearpage


\renewcommand*{\bibfont}{\footnotesize}
 
\bibliographystyle{IEEEtranN}

\end{document}